%% file: acm.tex
\documentclass[sigconf,nonacm]{acmart}

\AtBeginDocument{%
  }

\usepackage[utf8]{inputenc}
\usepackage[T1]{fontenc}

\usepackage{microtype}
\usepackage{cleveref}

\usepackage{graphicx}
\usepackage{float}
\usepackage{tabularray}
\usepackage{siunitx}
\usepackage{makecell}

\usepackage{algorithm}
\usepackage[noEnd=true, spaceRequire=false, endLComment=\ ]{algpseudocodex}

\usepackage{pgfplots}
\usepackage{tikz}

\usepackage{pifont}
\usepackage{comment}
\usepackage{enumitem}

\DeclareSIUnit{\fps}{fps}

\begin{document}

\title{Cross-Stage Sensorimotor Perception Scheduling
and Sparse Map Encoding for Efficient Edge Embodied Navigation}

\author{Yaotian Liu}
\affiliation{%
  \institution{Arizona State University}
  \city{Tempe}
  \state{Arizona}
  \country{USA}}
\email{yaotian_liu@asu.edu}

\author{Sri Sai Rakesh Nakkilla}
\affiliation{%
  \institution{Arizona State University}
  \city{Tempe}
  \state{Arizona}
  \country{USA}}
\email{snakkill@asu.edu}

\author{Xiangyu Zhou}
\affiliation{%
  \institution{Arizona State University}
  \city{Tempe}
  \state{Arizona}
  \country{USA}}
\email{xzhou185@asu.edu}

\author{Yu Cao}
\affiliation{%
  \institution{University of Minnesota}
  \city{Minneapolis}
  \state{Minnesota}
  \country{USA}}
\email{yucao@umn.edu}

\author{Jeff Zhang}
\affiliation{%
  \institution{Arizona State University}
  \city{Tempe}
  \state{Arizona}
  \country{USA}}
\email{jeffzhang@asu.edu}

\input{contents/abstract}

\maketitle

\input{contents/introduction}

\input{contents/background}

\input{contents/method}
\input{contents/experiments}
\input{contents/results}
\input{contents/conclusion}

\bibliographystyle{ACM-Reference-Format}
\bibliography{ref}

\end{document}

%% file: contents/abstract.tex
\begin{abstract}
Embodied agents must close a perception-to-action loop on embedded hardware under tight latency, memory, and energy budgets, making deployment a system-level co-design problem rather than a model-accuracy problem.
We study this challenge for modular Object Goal Navigation (ObjectNav), where our profiling shows semantic mapping dominates per-step latency while goal prediction dominates peak memory.
We formulate edge embodied navigation deployment as a budget-constrained design-space problem and introduce two orthogonal optimization knobs: \textbf{SKIP}, an adaptive sensorimotor scheduler that formalizes safe skipping as a bounded map-impact criterion and learns a lightweight predictor to estimate it from cheap sensor cues at each \texttt{FORWARD} step, exposing a principled quality-efficiency knob (depth-based updates are always retained); and \textbf{SCOUT}, a sparse-context encoder that couples submanifold sparse convolutions on active map regions with a lightweight dense context stream.
On HM3D across server and embedded platforms, SKIP+SCOUT delivers up to 1.7$\times$ end-to-end speedup, 50.5\% lower peak memory, and 7.1\% higher SPL than the dense baseline at the selected operating point, outperforming naively smaller perception backbones.
SKIP transfers to a second modular pipeline (PONI) with near-lossless performance and remains robust under depth-sensor noise.
Together, SKIP+SCOUT expose a family of device-aware Pareto operating points for edge physical AI systems.
\end{abstract}

%% file: contents/introduction.tex
\section{Introduction}
\label{sec:introduction}

\begin{figure*}[t]
    \centering
    \includegraphics[width=\textwidth]{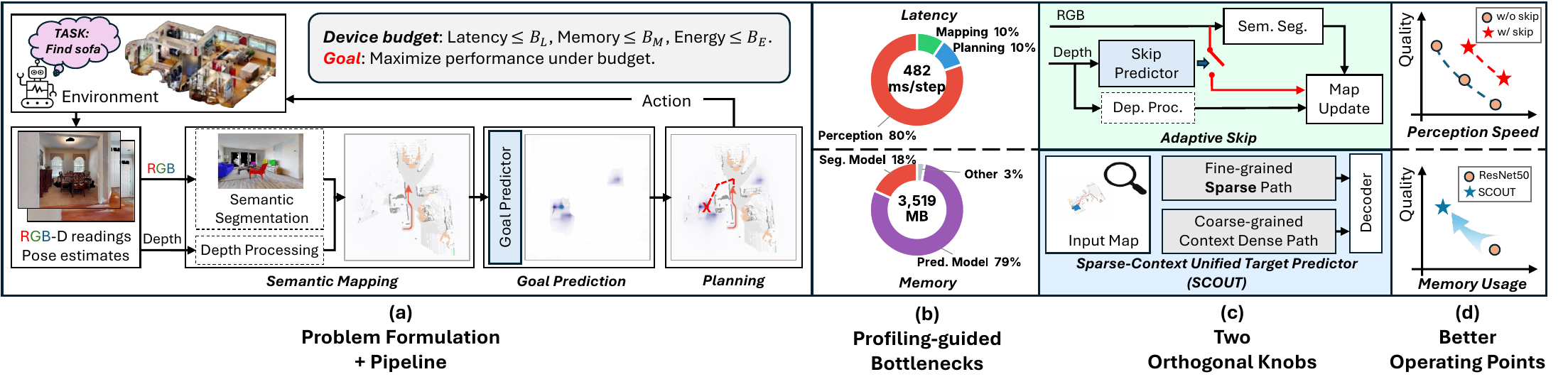}
    \caption{\textbf{Overview of the cross-stage design-space for edge embodied navigation.}
    (a)~An embodied agent closes a sensorimotor loop over RGB-D observations through three stages (semantic mapping, goal prediction, planning) under device-specific latency, memory, and energy budgets.
    (b)~Per-stage profiling on NVIDIA Jetson Orin reveals that semantic mapping dominates latency (83\%, 482\,ms/step) while the goal-prediction model dominates memory (79\% of 3{,}519\,MB).
    (c)~SKIP schedules \emph{when} the expensive semantic segmentation runs (depth-based map updates are always retained); SCOUT optimizes \emph{how} sparse maps are encoded for goal prediction. The two knobs are orthogonal.
    (d)~Together they expose a family of device-aware Pareto operating points over speed, memory, and energy that uniformly outperform dense baselines in our evaluation.}
    \label{fig:overview}
\end{figure*}

Physical AI systems such as household robots, mobile assistants, and inspection platforms must transform noisy RGB-D observations into actions within a tight sensorimotor loop on embedded compute~\cite{lin2018architectural, neuman2021robomorphic, boroujerdian2018mavbench}.
In these systems, the central challenge is not only the prediction accuracy, but whether the full perception-memory-planning stack can meet latency, memory, and energy budgets without sacrificing the task performance for real-world depolyment.

Object Goal Navigation (ObjectNav)~\cite{batra2020objectnav} provides a representative embodied benchmark for studying this problem: success depends on real-time perception, semantic spatial memory, and action selection in previously unseen environments.
Modern modular pipelines~\cite{chaplot2020object, ramakrishnan2022poni, zhai2023peanut} decompose this loop into three interpretable stages: (1)~\emph{semantic mapping}, which fuses segmented RGB-D frames into a top-down semantic map; (2)~\emph{goal prediction}, which infers likely target locations from the partial map; and (3)~\emph{planning}, which selects a goal and computes low-level actions.
Such modularity enables independent algorithmic advances, but it also exposes a system-level co-design challenge: each stage carries distinct latency and memory costs that interact under the tight resource budgets of edge platforms.

Deploying navigation pipelines on embedded hardware reveals significant bottlenecks.
Our profiling of a representative dense modular agent~\cite{zhai2023peanut} on NVIDIA Jetson Orin shows that \emph{semantic mapping accounts for 83\% of per-step latency} (482\,ms/step), while the \emph{goal-prediction model consumes 79\% of peak memory} (3{,}519\,MB total; see \Cref{fig:overview}b).
A straightforward solution is to shrink individual models of the embodied navigation pipeline, but our experiments show this is not the best remedy for  the edge deployment: smaller perception backbones reduce per-invocation cost while discarding semantic quality, however, they fail to exploit two structural properties within the embodied navigation setting, namely \emph{temporal redundancy across consecutive observations} and \emph{extreme sparsity in the accumulated semantic map}.

This motivates the central thesis of the paper: 
\emph{for edge embodied navigation, better deployment comes less from uniformly shrinking models and more from jointly exploiting temporal redundancy in perception and sparsity in semantic memory.}
We therefore cast edge embodied navigation as a budget-constrained design-space problem over \textbf{when} expensive perception should run and \textbf{how} sparse semantic maps should be encoded, and introduce two novel cross-stage optimization knobs: \textbf{SKIP} (adaptive perception scheduling) and \textbf{SCOUT} (sparse-context unified target predictor).
The two are orthogonal: SKIP reduces the frequency at which the perception stage is invoked without changing its per-invocation latency cost, while SCOUT reduces the per-invocation memory cost of the goal-prediction stage without changing its frequency.
Their interaction exposes a family of device-aware operating points for embodied navigation workloads across edge server and embedded platforms that neither knob reaches alone.

Our contributions are as follows:

    \textbf{1. Deployment formulation and design-space exploration.} We formulate edge embodied navigation deployment as a budget-constrained design-space over perception scheduling, sparse map encoding, and device configuration, subject to latency, memory, and energy constraints, and instantiate it through profiling-based device-aware configuration selection (\Cref{sec:background}).

    \textbf{2. SKIP: adaptive scheduling under bounded map-impact.} We formalize safe skipping as an $\ell_1$ map-impact criterion $L_m(t) < \tau$ on consecutive single-frame projections, learn a lightweight predictor $J_\theta$ that estimates this criterion from depth-histogram differences without invoking segmentation, and expose $(\tau, c)$ as a principled quality-efficiency knob; on HM3D, SKIP safely eliminates up to 88\% of forward-step segmentation invocations with only 3.6\% SPL degradation (\Cref{sec:skip}).

    \textbf{3. SCOUT: sparse-context map encoding.} We present SCOUT, a sparse-context goal-prediction encoder that combines submanifold sparse computation on active map cells with a lightweight dense context stream, reducing goal-prediction memory by 64\% and compute by 84\% relative to the dense baseline while improving SPL (\Cref{sec:scout}).

    \textbf{4. Device-aware design-space exploration.} We demonstrate device-specific Pareto operating points on edge server and embedded platforms, transfer of SKIP across a second modular pipeline (PONI~\cite{ramakrishnan2022poni}), and robustness to observation noise (\Cref{sec:results}).



%% file: contents/background.tex
\section{Background and Problem Formulation}
\label{sec:background}
We discuss the background and our problem formulation in this section.

\subsection{Modular ObjectNav Pipeline}
\label{sec:pipeline}

ObjectNav~\cite{batra2020objectnav, habitatchallenge2022} is an episodic task: the agent is initialized at a random pose in an unseen environment and must navigate to an instance of a given object category (e.g., ``chair'') using egocentric RGB-D observations and pose readings.
At each discrete time step the agent selects from four actions: \texttt{FORWARD} (\qty{0.25}{\m}), \texttt{TURN\_LEFT}/\texttt{TURN\_RIGHT} (\ang{30}), or \texttt{STOP}.
An episode succeeds if the agent calls \texttt{STOP} within \qty{1.0}{\m} of the target object, and each episode is limited to 500 steps. 

Approaches to ObjectNav can be broadly split into end-to-end reinforcement learning~\cite{wijmans2019dd, ramrakhya2022habitat, yadav2023ovrl, khandelwal2022simple} and modular pipelines that decompose perception, spatial memory, and planning into interpretable stages~\cite{chaplot2020object, ramakrishnan2022poni, zhai2023peanut, luo2022stubborn, liang2021sscnav, ye2021auxiliary}.
In this work, we focus on the modular navigation pipelines for two reasons: it currently leads end-to-end approaches on the HM3D ObjectNav Habitat Challenge~\cite{habitatchallenge2022}, and its interpretable stage decomposition decouples perception from policy, exposing a concrete design-space for device-aware optimization. 

Specifically, we decompose the modular navigation pipeline into three stages:

\paragraph{Stage 1: Semantic Mapping.}
RGB-D frames are segmented by a pre-trained model and projected into a top-down allocentric semantic map $\mathbf{m}_t \in \mathbb{R}^{(4+N)\times H\times W}$, where $N$ is the number of semantic categories and the first four channels encode obstacles, explored area, current agent location, and location history~\cite{chaplot2020object, zhai2023peanut}.
This stage is the most latency-intensive component (per step), dominated by the semantic segmentation backbone.

\paragraph{Stage 2: Goal Prediction.}
An encoder-decoder network ingests the partial semantic map $\mathbf{m}_t$ and outputs a probability map over likely target locations.
State-of-the-art Dense encoders such as ResNet-50 with PSPNet~\cite{zhao2017pyramid} achieve strong prediction quality but require hundreds of GFLOPs and over \qty{2}{\giga\byte} of activation memory, making them the dominant memory consumer in the pipeline~\cite{zhai2023peanut, ramakrishnan2022poni}.

\paragraph{Stage 3: Planning.}
The planning module merges goal selection and local replanning.
It selects a long-term goal from the predicted probability map, computes a shortest path via the Fast Marching Method~\cite{sethian1999fast}, and extracts a waypoint for the current step.
Replanning runs at every step, contributing a non-negligible fraction of per-step latency.
We do not directly optimize the planning algorithm in this work.

\subsection{Budget-Constrained Deployment Formulation}
\label{sec:formulation}

Deploying a modular ObjectNav agent on a target device~$d$ requires choosing a system configuration $c = (b_{\text{seg}}, s_{\text{skip}}, e_{\text{goal}}, r_{\text{deploy}})$, where $b_{\text{seg}}$ selects the segmentation backbone, $s_{\text{skip}}$ parameterizes the segmentation skip policy, $e_{\text{goal}}$ specifies the goal-prediction encoder architecture, and $r_{\text{deploy}}$ captures deployment choices (runtime precision, operator placement, platform mapping).

The deployment objective is $\max_{c}\; Q(c)$ subject to $L_d(c) \le B_L$, $M_d(c) \le B_M$, and $E_d(c) \le B_E$, where $Q(c)$ is navigation quality (measured primarily by SPL, success weighted by path length); $L_d(c)$, $M_d(c)$, and $E_d(c)$ are per-step latency, peak memory, and per-frame energy on device~$d$; and $(B_L, B_M, B_E)$ are device-specific budgets.

This formulation makes explicit that the design space is multi-dimensional.
Swapping to a smaller backbone changes $b_{\text{seg}}$ and reduces $L_d$ but may degrade $Q$ more than the latency gain warrants.
Adjusting the skip policy $s_{\text{skip}}$ reduces $L_d$ without changing $M_d$, while replacing the goal-prediction encoder $e_{\text{goal}}$ can dramatically reduce $M_d$ at modest $L_d$ cost.
The key insight is that these axes interact: a strong backbone combined with aggressive skipping can outperform a weak backbone run at every step, because the skip policy exploits temporal redundancy that the backbone choice cannot.
SKIP and SCOUT, introduced in \Cref{sec:method}, provide two principled cross-stage optimization knobs that operate on complementary dimensions of this configuration space.

\paragraph{Relation to conventional compression.}
Our focus on SKIP and SCOUT is deliberate: we study optimizations \emph{orthogonal} to conventional model compression.
Techniques like quantization~\cite{nagel2020adaround, jacob2018quantization, wang2019haq} and pruning~\cite{han2016deep, frankle2019lottery,li2017pruning} reduce per-invocation (memory, latency, and FLOPs) cost 
and are directly applicable to both the segmentation and goal-prediction stages; SKIP and SCOUT instead change \emph{when} expensive modules run and \emph{how} sparse intermediate maps are encoded.
The two families compose, and real-world deployments can combine them (e.g., int8 quantization on top of SKIP+SCOUT) for multiplicative gains.

%% file: contents/method.tex
\section{SKIP and SCOUT: Perception Scheduling and Sparse Map Encoding}
\label{sec:method}

We introduce two orthogonal optimization knobs that address complementary bottlenecks in the modular ObjectNav pipeline (\Cref{fig:overview}, panel~c).
SKIP (\Cref{sec:skip}) controls \emph{when} the expensive semantic segmentation executes, reducing per-step latency while always retaining depth-based map updates.
SCOUT (\Cref{sec:scout}) controls \emph{how} the sparse semantic map is encoded for goal prediction, reducing peak memory and computation.
\Cref{sec:config} describes how these knobs, together with backbone and deployment choices, define a design space of device-aware operating points.

\subsection{SKIP: Adaptive Perception Scheduling}
\label{sec:skip}

\paragraph{Task-aware perception scheduling under bounded frame-to-frame map impact.}
We frame adaptive perception as a scheduling problem with a principled bound on frame-to-frame map change.
Let $\mathbf{p}_t = [\mathbf{o}_t;\, \mathbf{s}_t]$ denote the single-frame projection into the global map frame, concatenating depth-derived channels $\mathbf{o}_t$ and the $N$ semantic channels $\mathbf{s}_t$ from the backbone.
We measure per-step \emph{map-impact} as the $\ell_1$ distance between consecutive projections,
\[
L_m(t) \;=\; \|\mathbf{p}_t - \mathbf{p}_{t-1}\|_1,
\]
and call step $t$ \emph{safe to skip} if $L_m(t) < \tau$: the current view is locally redundant with its predecessor, so re-running segmentation is unlikely to change the accumulated map $\mathbf{m}_t$.
Since measuring $L_m(t)$ requires running segmentation, SKIP learns a predictor $J_\theta$ that estimates $P(L_m(t) < \tau \mid \Delta\mathbf{h}_t)$ from a cheap sensor cue $\Delta\mathbf{h}_t$ (defined below) without invoking the backbone.
At inference, step $t$ is skipped iff $P(\text{safe}) > c$ for a confidence threshold $c$; the pair $(\tau, c)$ exposes a principled two-dimensional control over the quality--efficiency tradeoff, with $\tau$ the training-time map-impact budget and $c$ the inference-time predictor confidence.
Downstream navigation quality degrades when cumulative skipped map-impact grows too large; $\tau$ therefore acts as a direct per-step upper bound on how much novel information the scheduler may ignore. Empirically, $\tau < 200$ offers the best efficiency--quality tradeoff (\Cref{fig:skip_operating}). 

\paragraph{What is preserved.}
Depth-based channels of $\mathbf{m}_t$ (obstacle occupancy, explored-area tracking) are refreshed regardless of the skip decision, so only semantic layers lag and spatial awareness remains current.
Turn steps (\texttt{TURN\_LEFT}, \texttt{TURN\_RIGHT}) always trigger segmentation because the large viewpoint change produces large $L_m(t)$, violating the bounded map-impact assumption; the scheduler is therefore applied only on \texttt{FORWARD} steps.

\subsubsection{Skip Predictor}

The core of SKIP is a classifier $J_\theta$ that takes $\Delta \mathbf{h}_t = \mathbf{h}_t - \mathbf{h}_{t-1}$, the difference between histogrammed depth images of the current and previous steps ($n{=}50$ bins each), and predicts whether the map-impact $L_m(t)$ would fall below the threshold $\tau$ (\Cref{alg:skip}).
This single feature vector captures the degree of scene change between consecutive observations: large histogram shifts indicate new geometry (e.g., a doorway), while small shifts suggest redundant views during straight-line motion.

\paragraph{Training.}
Labels are generated offline: we run an exploration agent over 200 episodes from HM3D \emph{training} scenes, compute $L_m(t)$ at every step per the definition above, and assign the binary label $\mathbf{1}\!\left[L_m(t) < \tau\right]$.
The training and validation scene sets are disjoint, so no environment information leaks into evaluation.
We instantiate $J_\theta$ as a random-forest classifier~\cite{ho1995random} over the 50-dimensional histogram-difference feature; the regressor alternative is evaluated in \Cref{sec:results-classifier-vs-regressor}.

\paragraph{Inference.}
$J_\theta$ outputs $P(\text{safe} \mid \Delta \mathbf{h}_t)$; the step is skipped iff this probability exceeds $c$ (\Cref{alg:skip}).
Raising $c$ is conservative (fewer skips, lower false-skip risk); lowering $c$ trades semantic freshness for latency savings.
The predictor runs in approximately \qty{1.6}{\ms}, negligible relative to the $>$\qty{100}{\ms} cost of a full segmentation pass.

\begin{algorithm}[t]
\caption{SKIP inference at step $t$}\label{alg:skip}
\small
\begin{algorithmic}[1]
\If{$a_t \neq \texttt{FORWARD}$}
    \State Run segmentation \Comment{Turn steps: always segment}
\Else
    \State $\Delta \mathbf{h}_t \gets \mathbf{h}_t - \mathbf{h}_{t-1}$
    \If{$P(\text{safe} \mid \Delta \mathbf{h}_t) > c$}
        \State Skip segmentation; reuse previous labels
    \Else
        \State Run segmentation
    \EndIf
\EndIf
\State Always update depth-based map (obstacles, explored area)
\end{algorithmic}
\end{algorithm}

\begin{figure}[t]
    \centering
    \includegraphics[width=\columnwidth]{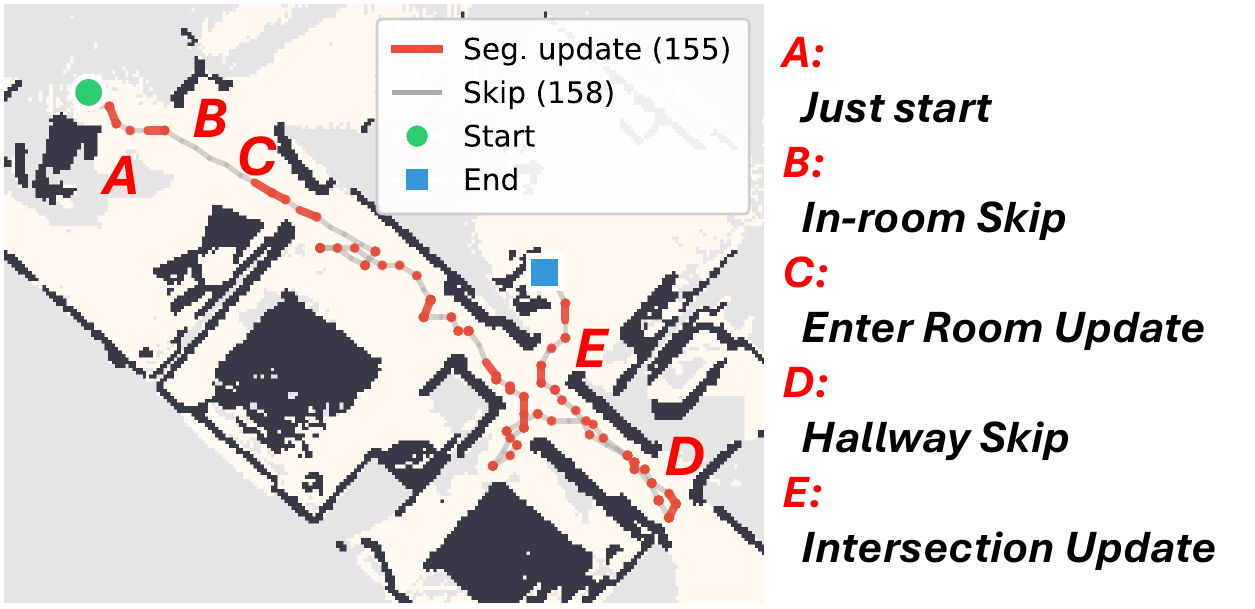}
    \caption{\textbf{SKIP decisions on a sample episode.} Red segments indicate steps where segmentation runs; gray segments are skipped. Updates concentrate at scene transitions (room entrances \textbf{C}, intersections \textbf{E}), while skips dominate during in-room motion (\textbf{B}) and hallway traversal (\textbf{D}).}
    \label{fig:skip_path}
\end{figure}

An alternative design would replace the classifier with a regressor that directly predicts the map-impact $L_m$, deferring the threshold comparison to inference time.
We evaluate both approaches in \Cref{sec:results-classifier-vs-regressor} and show that the classifier formulation achieves better skip accuracy for a given quality budget.

\subsubsection{Explainability Analysis}
\label{sec:skip-explain}

\begin{figure}[t]
    \centering
    \includegraphics[width=\columnwidth]{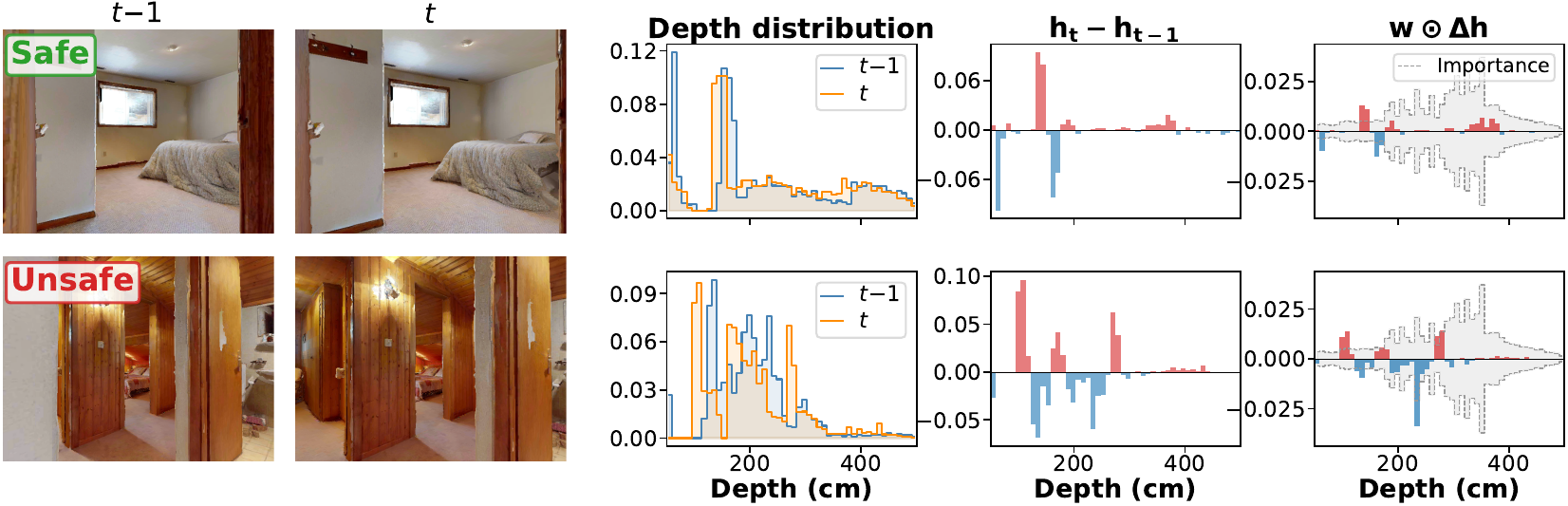}
    \caption{\textbf{SKIP explainability.} Two consecutive frames ($t{-}1$, $t$) for a safe-to-skip case (top) and an unsafe case (bottom). Left to right: RGB observations; depth histograms $\mathbf{h}_{t-1}$ and $\mathbf{h}_t$; their difference $\Delta \mathbf{h}_t = \mathbf{h}_t - \mathbf{h}_{t-1}$; and the importance-weighted contribution $\mathbf{w} \odot \Delta \mathbf{h}_t$ (gray dashed curve shows per-bin feature importance $\mathbf{w}$).
    In the safe case, consecutive frames are nearly identical and $\Delta \mathbf{h}_t$ is small, producing negligible weighted signal.
    In the unsafe case, the agent steps through a doorway, creating large shifts in the near-range bins (100--250\,cm) that align with high-importance features, triggering a full segmentation update.}
    \label{fig:skip_examples}
\end{figure}

We first show \emph{where} SKIP chooses to skip on a full episode (\Cref{fig:skip_path}), then zoom into individual frames to explain \emph{why} (\Cref{fig:skip_examples}).

\paragraph{Spatial pattern of skip decisions.}
\Cref{fig:skip_path} visualizes the skip/update decisions on a top-down map from a sample navigation episode.
Out of 313 steps, 158 are skipped (gray) and 155 trigger a full segmentation update (red).
The spatial pattern confirms the predictor's learned behavior:
segmentation updates cluster at the episode start where the scene is new (\textbf{A}), at room entrances where the depth distribution changes abruptly (\textbf{C}), and at hallway intersections where multiple paths diverge (\textbf{E}).
Skips dominate during in-room exploration where consecutive views overlap heavily (\textbf{B}) and along hallways with gradually changing depth (\textbf{D}).

\paragraph{Feature importance.}
To understand how SKIP processes individual frames, we visualize two representative cases in \Cref{fig:skip_examples}: a \emph{safe} skip (top row) and an \emph{unsafe} step that should not be skipped (bottom row).
We inspect the random-forest feature importances to identify which bins of the depth-histogram difference most influence skip decisions.
The predictor concentrates over half its importance (52\%) on mid-to-far depth bins (275--400\,cm), the range where changes most reliably signal scene transitions such as doorways or new rooms.
Near-range bins (50--150\,cm), which shift predictably with every forward step, and far-range bins (400--500\,cm), where depth readings are noisy, each receive roughly 10\% of total importance.

The predictor thus learns a general signal, \emph{concentrated depth redistribution in the mid-to-far range indicates new geometry}, rather than memorizing scenes, which underpins its transfer across environments and pipelines (\Cref{sec:results-transfer}).

\paragraph{Secondary effects and robustness.}
Two further properties of SKIP are analyzed empirically in the results: skipping redundant frames also acts as a semantic noise filter that improves SPL on hard episodes by avoiding occasionally noisy segmentation calls (\Cref{sec:results-noise-filter}), and the classifier's decision boundary remains stable under injected zero-mean depth-sensor noise (\Cref{sec:results-robustness}).

\subsection{SCOUT: Sparse Map Encoding for Goal Prediction}
\label{sec:scout}

The goal-prediction encoder is the dominant memory consumer in the pipeline (\Cref{sec:pipeline}).
The key observation is that the input semantic map $\mathbf{m}_t$ is highly sparse: the proportion of non-zero cells across $H \times W$ is typically below 10\%.
Dense encoders thus waste both computation and activation memory on empty space.

SCOUT exploits this sparsity through a dual-path architecture (\Cref{fig:scout_arch}):

\begin{figure}[t]
    \centering
    \includegraphics[width=\columnwidth]{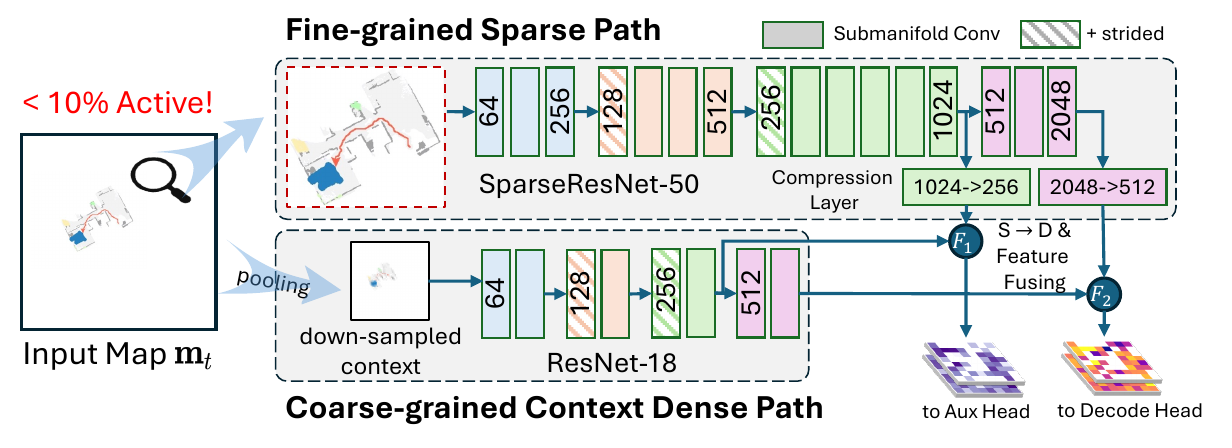}
    \caption{\textbf{SCOUT} (Sparse-Context Unified Target Predictor). The input semantic map $\mathbf{m}_t$ is less than 10\% active. The fine-grained sparse path (SparseResNet-50) uses submanifold convolutions and strided blocks to compute only on active cells, with $1{\times}1$ compression layers aligning channels. The coarse-grained context dense path (ResNet-18) processes a pooled, down-sampled version for global context. Features are fused via sparse-to-dense (S$\to$D) conversion at $F_1$ (to auxiliary head, train only) and $F_2$ (to decode head).}
    \label{fig:scout_arch}
\end{figure}

\paragraph{Sparse stream: adaptive zoom-in.}
We adopt SparseConvNet~\cite{graham2018sparseconvnet} to restrict computation and feature storage to the active cell set $\mathcal{A}_t$.
Submanifold sparse convolutions preserve the input support (no dilation), maintaining fine detail only where evidence exists and keeping the activation footprint proportional to $|\mathcal{A}_t|$ rather than $H{\times}W$.

\paragraph{Learned scale change via strided sparse convolution.}
To enlarge receptive fields without densifying the grid, SCOUT inserts occasional strided sparse blocks that downsample the active set.
Each stride acts as a learned zoom step, aggregating nearby evidence and reducing the number of stored features for deeper layers.

\paragraph{Dense context stream.}
A lightweight ResNet-18 runs in parallel at reduced resolution to supply global context.
After $1{\times}1$ channel compression and sparse-to-dense conversion, sparse (high detail, zoomed-in) and dense (global, zoomed-out) features are fused at two points ($F_1$, $F_2$) before decoding via a PSPNet~\cite{zhao2017pyramid} head.
A train-only auxiliary head branches from~$F_1$.

\paragraph{Why the interaction matters.}
Neither the sparse stream nor the dense stream alone is sufficient.
Sparse-only processing lacks long-range context, while dense-only processing wastes resources on empty space.
The positive interaction between sparse fine-grained detail and lightweight dense context is the architectural insight behind SCOUT's quality-cost tradeoff (\Cref{sec:results-scout}).

\paragraph{Training.}
SCOUT follows prior work~\cite{zhai2023peanut} for training data generation and loss function.
We use the PSPNet architecture with SCOUT as the backbone, an auxiliary loss weight of 0.4, the Adam optimizer ($\alpha{=}0.0005$, $\beta_1{=}0.9$, $\beta_2{=}0.999$), batch size~8, and a ``poly'' learning-rate schedule~\cite{zhao2017pyramid}.

\subsection{Device-Aware Configuration Selection}
\label{sec:config}

SKIP and SCOUT together define a joint configuration space over three tunable axes: the segmentation backbone~$b_{\text{seg}}$ (e.g., Mask~R-CNN, ESANet, SegFormer), the skip policy~$s_{\text{skip}}$ (threshold~$\tau$ and confidence gate~$c$), and the goal-prediction encoder~$e_{\text{goal}}$ (PSPNet, FPN, HRNet, SCOUT).
Each axis affects latency and memory differently:
$b_{\text{seg}}$ controls segmentation cost per invocation;
$s_{\text{skip}}$ reduces how often that cost is paid;
$e_{\text{goal}}$ determines goal-prediction compute and, critically, peak memory.

\paragraph{Selection procedure.}
For a target device with budget $(B_L, B_M, B_E)$, we profile per-step latency $L_d(c)$, peak memory $M_d(c)$, and per-frame energy $E_d(c)$ for each configuration~$c$ in the discrete design space, discard configurations that violate any constraint, and select the feasible configuration maximizing SPL per the formulation in \Cref{sec:formulation}.
Because profiling is offline and the predictor retrains in seconds on a single CPU, the search cost is negligible.

\paragraph{Concrete example.}
As budgets shrink, the selection automatically falls back from Mask~R-CNN + SCOUT + SKIP on a server to ESANet + SCOUT + SKIP on a Jetson Orin with a tight latency budget, yielding a \emph{family of device-specific operating points along the Pareto frontier}; exact configurations and numbers appear in \Cref{sec:results-pareto,sec:results-final}.

%% file: contents/experiments.tex
\section{Experimental Methodology}
\label{sec:experiments}

This section discusses our experimental setup.

\subsection{Datasets and Simulation}

We evaluate on the Habitat-Matterport 3D (HM3D) dataset~\cite{ramakrishnan2021habitatmatterport} using the Habitat simulator~\cite{savva2019habitat} under the standard ObjectNav challenge setting~\cite{habitatchallenge2022} with six goal categories.
The training set contains 3{,}971{,}566 episodes across 800 scenes; the validation (\texttt{VAL}) set contains 2{,}000 episodes across 100 scenes.
We also report on the hidden \texttt{TEST-STANDARD} split via the online evaluation server.
The agent receives $640{\times}480$ RGB-D observations with a \ang{79} horizontal field of view.

For the SKIP transferability study (\Cref{sec:results-transfer}), we additionally evaluate on the Gibson dataset~\cite{xia2018gibson} using the PONI~\cite{ramakrishnan2022poni} pipeline, a second modular ObjectNav system with a different segmentation backbone and goal-prediction architecture.

\subsection{Modular Pipelines}

Our primary instantiation uses the PEANUT~\cite{zhai2023peanut} pipeline as the dense baseline, comprising Mask~R-CNN (ResNet-101)~\cite{he2017mask} for semantic segmentation and ResNet-50 + PSPNet~\cite{zhao2017pyramid} for goal prediction.
To demonstrate generality, we also apply SKIP to PONI~\cite{ramakrishnan2022poni}, which uses a different segmentation model and goal encoder.

\subsection{Hardware Platforms}
\label{sec:platforms}

We measure latency and peak memory on two platforms spanning the edge server to embedded system spectrum:
\begin{enumerate}[leftmargin=2em]
    \item \textbf{Edge Server}: AMD 3960X CPU + NVIDIA RTX A6000 GPU.
    \item \textbf{NVIDIA Jetson Orin}: Ampere GPU + 8-core Arm Cortex-A78AE, \qty{16}{\giga\byte} shared RAM.
\end{enumerate}
All timing measurements use wall-clock time averaged over complete episodes.
GPU memory footprint is reported as the peak allocation by the PyTorch caching allocator during steady-state inference (post-warmup), ensuring transient allocations during warmup do not inflate the reported numbers.
Energy on Jetson Orin NX is measured by polling the on-board INA3221 power rails via \texttt{tegrastats} at 200\,ms intervals: per-stage power is the idle-subtracted mean of VDD\_IN over a 25\,s steady-state window (preceded by a 15\,s idle baseline and 30\,s warm-up), and per-frame energy equals this net power divided by measured throughput.

\subsection{Metrics}

To messure the naviation quality, we adopt the standard ObjectNav metrics:
\begin{itemize}[leftmargin=1.5em]
    \item \textbf{SPL} (Success weighted by Path Length)~\cite{batra2020objectnav}: our primary quality metric, jointly capturing success rate and path efficiency.
    \item \textbf{SoftSPL} (S-SPL)~\cite{habitatchallenge2022}: a softer variant that tracks progress even in failed episodes; our secondary quality metric.
    \item \textbf{Success Rate} (SR): reported for context only; not used as the primary comparison axis.
\end{itemize}

For hardware efficiency, we report the following:
\begin{itemize}[leftmargin=1.5em]
    \item \textbf{Skip ratio}: fraction of steps where semantic segmentation is bypassed (depth-based map updates still execute).
    \item \textbf{GFLOPs}: floating-point operations for the goal-prediction encoder.
    \item \textbf{Per-step latency} (ms/step) and \textbf{peak memory} (MB) on each platform. We also report \textbf{energy consumption} (J/frame) on NVIDIA Jetson Orin.

\end{itemize}

\subsection{Measurement Methodology}

Design-space exploration for SKIP operating points (varying the skip threshold~$\tau$) uses a 200-episode subset of HM3D \texttt{VAL} to keep the search tractable.
Final reported numbers use the full 2{,}000-episode \texttt{VAL} split.
The SKIP predictor is a random-forest classifier trained on depth-histogram differences ($\Delta \mathbf{h}_t = \mathbf{h}_t - \mathbf{h}_{t-1}$, $n{=}50$ bins) from 200 HM3D training episodes whose scenes are disjoint from the validation and test scenes, ensuring no environment information leakage.
SCOUT is trained following the protocol of~\cite{zhai2023peanut}.

%% file: contents/results.tex
\section{Results}
\label{sec:results}

We organize the evaluation around the central thesis: \emph{better edge embodied navigation comes less from uniformly shrinking models and more from jointly exploiting temporal redundancy in perception (SKIP) and sparsity in semantic memory (SCOUT)}, substantiated below for scheduling (\Cref{sec:results-scheduling}), encoding (\Cref{sec:results-scout}), and end-to-end Pareto operating points (\Cref{sec:results-pareto}).

\subsection{Why a Learned SKIP Scheduler Beats Naive Alternatives}
\label{sec:results-scheduling}

Reducing per-step latency could be achieved via two naive alternatives: (i) replace the segmentation backbone with a smaller model, or (ii) schedule the existing backbone with a hand-designed heuristic on the same feature (histogrammed depth) our predictor consumes. We show both degrade navigation quality substantially more than our learned scheduler at comparable efficiency gains.

\paragraph{Naive model shrinking.}
We select two representative lightweight alternatives, ESANet (ResNet-34)~\cite{seichter2021efficient} and SegFormer~B0~\cite{xie2021segformer}, and retrain them under the same protocol as the default Mask~R-CNN (ResNet-101) to ensure a fair comparison.
\Cref{tab:seg_backbone} reports the results: smaller backbones reduce latency but incur disproportionate quality loss.

\begin{table}[t]
\centering
\caption{Scheduling vs.\ naive model shrinking on PEANUT/HM3D. Smaller backbones trade quality for speed; SKIP retains the strong Mask~R-CNN backbone and achieves comparable or better segmentation-stage speedups with far less quality loss. Seg.\ Speedup is the relative speedup of the segmentation stage only: backbone latency ratio for swaps, $1/(1 - \text{amortized skip ratio})$ for SKIP.}
\label{tab:seg_backbone}
\begin{tabular}{lcccc}
\toprule
Model & Succ.\ & SPL & S-SPL & Seg.\ Speedup \\
\midrule
\multicolumn{5}{l}{\emph{Backbone swap (retrained, 2{,}000 ep)}} \\
Mask R-CNN R-101 & 0.585 & 0.308 & 0.339 & 1.0$\times$ \\
ESANet R-34      & 0.559 & 0.276 & 0.314 & 2.2$\times$ \\
SegFormer B0     & 0.519 & 0.233 & 0.267 & 3.4$\times$ \\
\midrule
\multicolumn{5}{l}{\emph{Mask R-CNN + SKIP (2{,}000 ep)}} \\
$\tau{=}75,\; c{=}0.50$  & 0.599 & 0.306 & 0.337 & 1.2$\times$ \\
$\tau{=}150,\; c{=}0.50$ & 0.595 & 0.301 & 0.335 & 1.5$\times$ \\
$\tau{=}200,\; c{=}0.50$ & 0.595 & 0.297 & 0.328 & 1.9$\times$ \\
\bottomrule
\end{tabular}
\end{table}

\begin{figure}[t]
    \centering
    \includegraphics[width=\columnwidth]{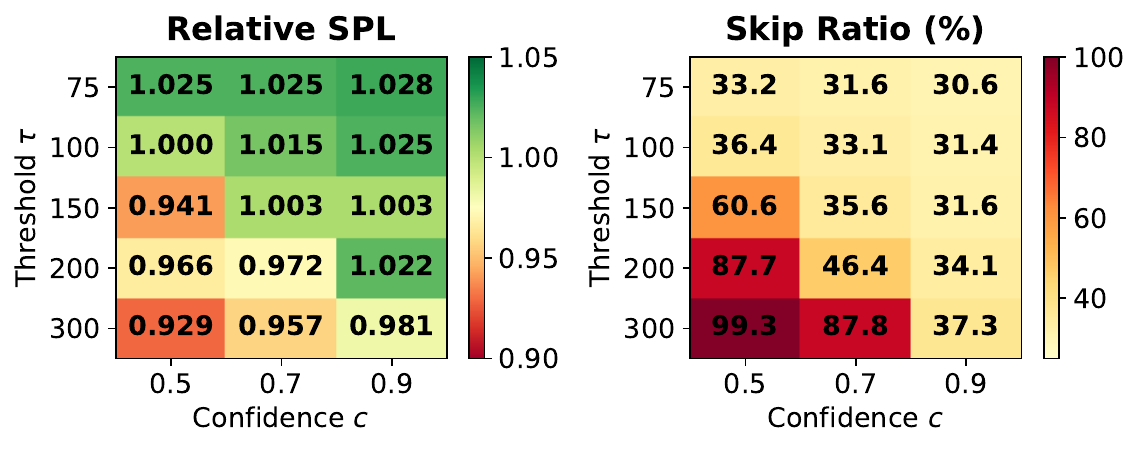}
    \caption{SKIP operating points on PEANUT/HM3D (200 episodes). Left: relative SPL (green = above baseline). Right: skip ratio of forward steps. The threshold~$\tau$ and confidence gate~$c$ jointly control the quality-latency tradeoff.}
    \label{fig:skip_operating}
\end{figure}

\paragraph{Heuristic scheduling.}
We evaluate three natural heuristic baselines: periodic skipping every $k$-th forward step (content-blind), and $\ell_1$ norm / Jensen-Shannon divergence~\cite{MENENDEZ1997307} thresholds on $\Delta\mathbf{h}_t$ (content-aware, same feature our predictor uses) (\Cref{tab:skip_heuristic}).
At matched ${\sim}90\%$ skip, all three lose 0.059--0.071 SPL (19--23\% drop), 5--6.5$\times$ worse than our learned SKIP's 0.011 loss at 87.7\% skip (\Cref{tab:seg_backbone}, \Cref{fig:skip_operating}).
Content-blind periodic skipping is strictly worse than the statistical thresholds, and both thresholds weight every bin equally; the random forest instead concentrates importance on mid-range bins that signal scene transitions (\Cref{sec:skip-explain}).

\begin{table}[t]
\centering
\caption{Heuristic skip scheduling baselines on PEANUT/HM3D (200 ep, baseline SPL\,=\,0.308). All three heuristics lose 13--27\% SPL across their sweeps; see \Cref{tab:seg_backbone} and \Cref{fig:skip_operating} for our learned SKIP at matched skip ratios.}
\label{tab:skip_heuristic}
\small
\begin{tabular}{lcccc}
\toprule
Method & Threshold & Succ.\ & SPL & Skip\% \\
\midrule
\textbf{Periodic $k$} & $k{=}2 / 5$           & .425 / .485 & .225 / .237 & 73.5 / 89.2 \\
\textbf{$\ell_1$}     & $\gamma{=}0.80 / 1.10$ & .520 / .475 & .267 / .243 & 76.8 / 89.7 \\
\textbf{JS}           & $\gamma{=}0.13 / 0.28$ & .505 / .505 & .252 / .249 & 71.3 / 90.2 \\
\bottomrule
\end{tabular}
\end{table}

\paragraph{Learned SKIP}
Our predictor avoids both pitfalls: it retains the strong Mask~R-CNN backbone and schedules it with a random forest on $\Delta\mathbf{h}_t$ with \emph{learned} bin weighting.
\Cref{fig:skip_operating} sweeps the full operating space over $\tau$ and~$c$: skip ratios span 31\%--88\% with relative SPL from 1.028 (conservative) to 0.964 (aggressive).

\textbf{The key message}: \emph{a learned scheduler dominates both naive model shrinking and fixed-threshold heuristics}, trading only 3.6\% SPL for 88\% skip versus 10--24\% loss for naive shrinking at matched speedups and 19--23\% for heuristics at matched skip.

\subsection{Classifier vs.\ Regressor for Skip Prediction}
\label{sec:results-classifier-vs-regressor}

Another alternative is a regressor that predicts the continuous map-impact $L_m$ and thresholds at inference. On Gibson (200 ep, baseline SPL\,=\,0.351), the regressor degrades SPL by 3--5$\times$ more than the classifier across thresholds (up to $\Delta$SPL\,=\,$-$0.043 vs.\ $-$0.008) because it suffers from regression-to-the-mean near the decision boundary; the classifier also exposes the confidence gate~$c$ as a second tuning knob (\Cref{alg:skip}).

\subsection{SKIP Robustness to Sensor Noise}
\label{sec:results-robustness}

Real depth sensors introduce per-pixel noise that can shift the histogram-difference feature and, in principle, destabilize SKIP's decision boundary.
To quantify this, we inject zero-mean Gaussian noise at increasing $\sigma$ into depth observations and measure both decision-boundary stability (20{,}862 held-out samples, fraction of skip/no-skip decisions that match the noise-free baseline) and end-to-end SPL (Gibson, 250 episodes).
At $\sigma{=}5$\,cm (typical structured-light), decision agreement with the noise-free baseline exceeds 92\% and SPL drops only from 0.365 to 0.360; even at $\sigma{=}20$\,cm (heavily degraded), decision agreement stays above 87\% and SPL holds at 0.359.
Histogramming aggregates ${\sim}3{\times}10^5$ pixels into 50 bins, absorbing per-pixel noise well below the classifier's decision boundary.
This study targets zero-mean Gaussian depth noise; systematic perturbations such as missing-depth dropout, reflective-surface holes, and sensor bias or drift are out of scope and a direction for future work.

\subsection{SKIP as a Semantic Noise Filter}
\label{sec:results-noise-filter}

An unexpected finding is that SKIP does not merely preserve navigation quality; it can actively \emph{improve} the quality in difficult episodes.
\Cref{tab:noise_filter} stratifies episodes by baseline SPL (without skipping) and measures the per-stratum effect of conservative ($\tau{=}75$) and aggressive ($\tau{=}200$) SKIP on PEANUT/HM3D \texttt{VAL}, restricted to episodes with successful baseline completions; failed baseline runs (SPL\,=\,0) are excluded because their per-episode $\Delta$SPL is not a meaningful comparison.

\begin{table}[t]
\centering
\caption{SKIP effect stratified by episode difficulty (baseline SPL) across episodes with successful baseline completions from PEANUT/HM3D \texttt{VAL}. SKIP improves SPL in hard episodes (low baseline) by filtering semantic noise, at a small cost in easy episodes.}
\label{tab:noise_filter}
\begin{tabular}{lccc}
\toprule
Baseline SPL & Episodes (\%) & $\Delta$SPL ($\tau{=}75$) & $\Delta$SPL ($\tau{=}200$) \\
\midrule
$[0.00, 0.15)$ & 3.3  & $+$0.071 & $+$0.144 \\
$[0.15, 0.30)$ & 12.6 & $+$0.039 & $+$0.063 \\
$[0.30, 0.50)$ & 29.8 & $+$0.017 & $+$0.032 \\
$[0.50, 0.70)$ & 30.7 & $-$0.019 & $-$0.052 \\
$[0.70, 1.00)$ & 23.5 & $-$0.025 & $-$0.078 \\
\bottomrule
\end{tabular}
\end{table}

The effect is strongly asymmetric.
In hard episodes (baseline SPL $<$ 0.30) the agent wanders longer, accumulating more false detections that corrupt the semantic map; skipping redundant frames avoids this injection, gaining up to +0.144 SPL.
In easy episodes (baseline SPL $>$ 0.50) the path is already near-optimal, and occasionally missing a genuine detection costs SPL.
At $\tau{=}75$ the two effects roughly cancel; at $\tau{=}200$ both grow.
This reframes SKIP beyond pure efficiency: \emph{it acts as a semantic noise filter, improving path efficiency in hard episodes at a small cost in easy ones.}

\subsection{SKIP Transferability}
\label{sec:results-transfer}

To show SKIP is a reusable knob rather than workload-specific, we apply it unchanged to PONI~\cite{ramakrishnan2022poni} on Gibson, a different modular pipeline with a different segmentation backbone and goal encoder.
At $(\tau{=}300, c{=}0.70)$ on the full 1{,}005-episode set, SKIP eliminates 67.1\% of segmentation calls with SPL preserved (0.353 vs.\ 0.351 baseline; Success 0.675 vs.\ 0.672), providing the evidence that SKIP transfers to other modular pipelines without any architectural changes.
The main failure mode is a delayed map update at abrupt scene transitions, which the planner compensates for within a few steps; SPL degradation correlates with $\tau$ and is negligible at the selected operating point.

\subsection{SCOUT Architecture Tradeoffs}
\label{sec:results-scout}

We evaluate SCOUT through a sparse-dense interaction ablation and an efficiency comparison against alternative encoder architectures for goal prediction, all trained under identical conditions.
\Cref{tab:scout_arch} isolates the contribution of each path; \Cref{tab:scout_efficiency} compares efficiency across architectures.
We deliberately include FPN~\cite{lin2017feature} and HRNet~\cite{wang2020deep} as baselines because both are established hierarchical, multi-resolution architectures that fuse features across scales, the same design principle underlying SCOUT.
Comparing against them allows us separate the benefit of multi-resolution fusion itself from SCOUT's specific sparse-dense decomposition.

\begin{table}[t]
\centering
\caption{Sparse-dense interaction ablation for goal-prediction encoder (SPL, 200-episode subset). SCOUT combines a sparse R50 primary stream with a lightweight dense R18 context stream; neither path alone matches the combined encoder.}
\label{tab:scout_arch}
\begin{tabular}{lcc}
\toprule
 & Single-path & Two-path (+ R18 context) \\
\midrule
Dense R50  & 0.308 (PSPNet R50)  & 0.281 (Dense SCOUT) \\
Sparse R50 & 0.252 (Sparse R50)  & \textbf{0.361} (SCOUT) \\
\bottomrule
\end{tabular}
\end{table}

\begin{table}[t]
\centering
\caption{Efficiency comparison of goal-prediction encoders (SPL on full 2{,}000-episode validation). SCOUT uses 6.2$\times$ fewer FLOPs than PSPNet R50 and 2.8$\times$ less memory while achieving the highest SPL.}
\label{tab:scout_efficiency}
\small
\begin{tabular}{lcccc}
\toprule
Model & GFLOPs & Mem.\,(MB) & Time\,(ms) & SPL \\
\midrule
\textbf{SCOUT} & \textbf{${\sim}$32} & \textbf{1006} & \textbf{27} & \textbf{0.333} \\
FPN          & 34.5  & 1059 & 14 & 0.322 \\
HRNet        & 35.3  & 2038 & 57 & 0.317 \\
PSPNet R50   & 200.8 & 2782 & 43 & 0.308 \\
\bottomrule
\end{tabular}
\end{table}

Three key findings emerge:
\begin{enumerate}[leftmargin=2em]
    \item \textbf{Sparse-only is insufficient.} A standalone Sparse R50 achieves SPL\,=\,0.252, lacking the long-range spatial context needed for effective goal prediction.
    \item \textbf{Dense-only is wasteful.} PSPNet R50 reaches SPL\,=\,0.308 but at 200.8\,GFLOPs and 2782\,MB; Dense SCOUT adds a context stream yet degrades to 0.281, spending resources on empty map cells.
    \item \textbf{The sparse+dense interaction is the key.} SCOUT couples sparse fine-grained detail with lightweight dense R18 context, achieving the best SPL in the ablation (0.361) and on the full validation set (0.333, \Cref{tab:scout_efficiency}), at 6.2$\times$ fewer FLOPs and 2.8$\times$ less memory than PSPNet R50.
\end{enumerate}

FPN (0.322) and HRNet (0.317) both improve over PSPNet R50 (0.308), confirming that multi-scale fusion helps on sparse semantic maps; SCOUT's sparse-dense decomposition goes further, reaching SPL 0.333 at 1{,}006\,MB (half HRNet's memory, tied with FPN).
Although FPN is faster per invocation (14 vs.\ 27\,ms), goal prediction runs only every 10 steps, so the amortized latency cost is just 1.3\,ms/step; SCOUT is therefore the best quality-memory operating point for goal prediction.

\subsection{End-to-End Pareto Fronts}
\label{sec:results-pareto}

\begin{figure}[t]
    \centering
    \includegraphics[width=\columnwidth]{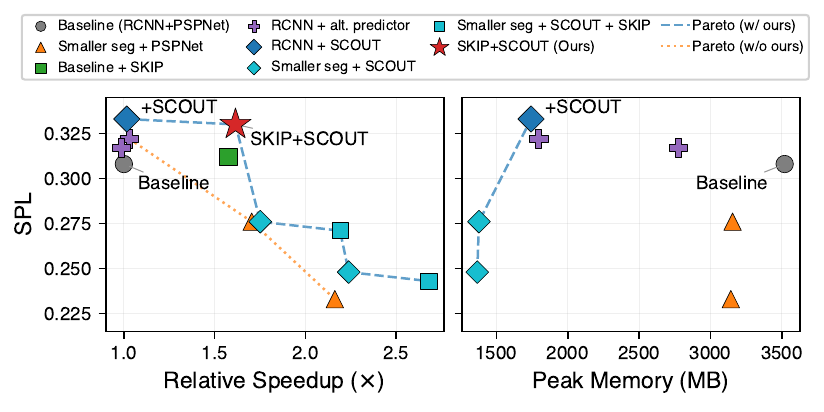}
    \caption{Pareto frontier across system configurations. Left: SPL vs.\ relative speedup. Right: SPL vs.\ peak memory. Blue dashed line: Pareto front with SKIP/SCOUT; orange dotted line: Pareto front without. All points on the ``with ours'' front use SCOUT.}
    \label{fig:pareto}
\end{figure}

We instantiate the configuration space of \Cref{sec:formulation} by sweeping three axes:
\begin{align*}
b_{\text{seg}} &\in \{\text{Mask~R-CNN}, \text{ESANet}, \text{SegFormer}\}, \\
e_{\text{goal}} &\in \{\text{PSPNet}, \text{FPN}, \text{HRNet}, \text{SCOUT}\}, \\
s_{\text{skip}} &\in \{\text{off},\; \text{on (per-backbone operating point of \Cref{sec:results-scheduling})}\},
\end{align*}
yielding 12 configurations profiled on both platforms.
\Cref{fig:pareto} plots SPL against relative speedup (left) and peak memory (right), with Pareto fronts for configurations with and without our techniques.

Three observations stand out.
First, SCOUT alone shifts the frontier: replacing PSPNet R50 with SCOUT at the same segmentation backbone improves SPL by 8.1\% (0.308$\to$0.333) while cutting peak memory by 50.5\% (3{,}519$\to$1{,}743\,MB) at comparable latency.
Second, SKIP+SCOUT adds speedup on top: a 1.6$\times$ speedup over the baseline with only 0.003 SPL below SCOUT-only (0.330 vs.\ 0.333), and still 7.1\% above the baseline.
Third, naive model shrinking is dominated: ESANet+PSPNet reaches 1.7$\times$ speedup but drops SPL by 10.4\%; SegFormer+PSPNet reaches 2.2$\times$ but loses 24.4\%.
Across evaluated configurations, the ``with ours'' Pareto front lies uniformly above the ``without'' front, showing that SKIP and SCOUT provide better quality-efficiency tradeoffs than backbone swapping alone.

\paragraph{Cross-platform impact.}
On our edge server (AMD 3960X + A6000), SKIP+SCOUT reduces per-step latency from 101.4\,ms to 62.8\,ms (9.9$\to$15.9\,FPS, 1.6$\times$).
On NVIDIA Jetson Orin, where the weaker GPU makes segmentation a larger fraction of total latency, the speedup is even greater: 482\,ms$\to$283\,ms (2.1$\to$3.5\,FPS, 1.7$\times$).
Peak memory drops from 3{,}519\,MB to 1{,}743\,MB on both platforms.
\Cref{tab:orin_breakdown} shows the stage-wise latency and energy decomposition on Orin: most of the savings come from SKIP eliminating 48\% of segmentation invocations, while SCOUT contributes a $2.9\times$ reduction in goal-prediction energy.
Per-frame energy drops from 7.93\,J to 4.24\,J, at a 47\% reduction.

\begin{table}[t]
\centering
\caption{Stage-wise latency and energy breakdown on  Jetson Orin. Total per-step energy drops by 47\% under SKIP+SCOUT, dominated by the segmentation savings from SKIP.}
\label{tab:orin_breakdown}
\small
\begin{tabular}{lcccc}
\toprule
 & \multicolumn{2}{c}{Baseline (RCNN+PSPNet)} & \multicolumn{2}{c}{SKIP+SCOUT} \\
\cmidrule(lr){2-3} \cmidrule(lr){4-5}
Stage & ms & J/frame & ms & J/frame \\
\midrule
Segmentation      & 372  & 6.55 & 193 & 3.41 \\
Goal pred.\ (/10) & 49.6 & 0.89 & 25.6 & 0.31 \\
Mapping           & 30   & 0.26 & 30 & 0.26 \\
Planning          & 30.7 & 0.23 & 30.7 & 0.23 \\
SKIP check        & --   & --   & 3.3 & 0.025 \\
\midrule
\textbf{Total}    & \textbf{482} & \textbf{7.93} & \textbf{283} & \textbf{4.24} \\
\bottomrule
\end{tabular}
\end{table}

\paragraph{SPL-adjusted energy (proxy).}
Because a higher-SPL agent reaches the goal in fewer steps, per-frame energy understates the deployment benefit.
As a first-order surrogate, $E_{\text{adj}}{\propto}E_{\text{frame}}/\text{SPL}$ estimates an $\sim\!50\%$ per-episode energy reduction on Orin (vs.\ 47\% per-frame); SPL mixes success and path efficiency, so the true saving lies between these two figures.

\subsection{Final Navigation Quality}
\label{sec:results-final}

\Cref{tab:main_results} reports end-to-end navigation quality on both \texttt{VAL} and hidden \texttt{TEST-STANDARD} splits, benchmarked against published end-to-end and modular methods.

\begin{table}[t]
\centering
\caption{ObjectNav results on HM3D \texttt{VAL} and \texttt{TEST-STANDARD}. SKIP+SCOUT achieves competitive navigation quality while providing substantially better efficiency (see \Cref{fig:pareto}).}
\label{tab:main_results}
\small
\begin{tabular}{lccccccc}
\toprule
 & \multicolumn{3}{c}{\texttt{VAL}} & & \multicolumn{3}{c}{\texttt{TEST-STANDARD}} \\
\cmidrule{2-4} \cmidrule{6-8}
Method & SPL & S-SPL & SR & & SPL & S-SPL & SR \\
\midrule
DD-PPO~\cite{wijmans2019dd} & 14.2 & -- & 27.9 & & 12.0 & 22.0 & 26.0 \\
Habitat-Web~\cite{ramrakhya2022habitat} & 23.8 & -- & 57.6 & & 22.0 & 26.0 & 55.0 \\
OVRL-V2~\cite{yadav2023ovrl} & 28.1 & -- & 64.7 & & 29.0 & -- & 64.0 \\
ProcTHOR~\cite{deitke2022procthor} & -- & -- & -- & & 32.0 & 38.0 & 54.0 \\
PEANUT~\cite{zhai2023peanut} & 30.8 & 33.9 & 60.6 & & 33.0 & 36.0 & 64.0 \\
SCOUT (w/o SKIP) & \textbf{33.3} & \textbf{36.6} & 60.5 & & 34.2 & 37.8 & 60.8 \\
\textbf{SKIP+SCOUT} & 33.0 & 36.2 & 60.4 & & \textbf{34.3} & \textbf{38.1} & 60.8 \\
\bottomrule
\end{tabular}
\end{table}

On \texttt{VAL} (2{,}000 episodes), SCOUT without SKIP achieves the highest absolute SPL (33.3) and SoftSPL (36.6), improving over the dense baseline (PEANUT: 30.8 / 33.9) by 8.1\% and 8.0\%.
Adding SKIP trades 0.3 SPL for a 1.6$\times$ speedup and 50.5\% memory reduction, making SKIP+SCOUT (SPL\,=\,33.0) the best quality-speed-memory operating point.
On the hidden \texttt{TEST-STANDARD} split, SKIP+SCOUT reaches 34.3 SPL and 38.1 SoftSPL.

These results should be interpreted through the lens of the budget-constrained formulation (\Cref{sec:formulation}): the contribution is not a single operating point on a leaderboard, but a family of Pareto-optimal configurations that a deployment engineer can select from given device-specific budgets $(B_L, B_M, B_E)$.
On our server ($B_L{=}120$\,ms, $B_M{=}4$\,GB), the procedure selects Mask~R-CNN + SCOUT + SKIP (SPL\,=\,0.330, 62.8\,ms, 1{,}743\,MB).
On Jetson Orin under a tight latency budget ($B_L{=}200$\,ms, $B_M{=}2$\,GB), the selection falls back to ESANet + SCOUT + SKIP (SPL\,=\,0.271, 177.1\,ms, 1{,}378\,MB).
In both cases, the selected configuration lies on the Pareto front.

%% file: contents/conclusion.tex
\section{Conclusion}
\label{sec:conclusion}

We presented a cross-stage design-space methodology and two optimization knobs, SKIP and SCOUT, for deploying modular embodied-navigation agents under latency, memory, and energy budgets on edge devices.
Our experiments support the central thesis: better deployment comes less from uniformly shrinking models and more from jointly exploiting temporal redundancy in perception (SKIP) and sparsity in semantic memory (SCOUT), delivering up to 1.7$\times$ speedup, 50.5\% memory reduction, and 7.1\% higher SPL over the dense baseline.
We also show that SKIP successfully transfers to other modular navigation pipelines with near-lossless performance, and the proposed methodology generalizes to other multi-stage perception systems where expensive modules can be scheduled and sparse intermediate representations exploited.